\documentclass{article}
\usepackage{spconf,amsmath,graphicx}
\usepackage{arydshln}
\usepackage{cite}
\usepackage[square, comma, sort&compress, numbers]{natbib}
\usepackage{epsfig}
\usepackage{epstopdf}
\usepackage{citesort}
\usepackage{url}
\usepackage{array}
\usepackage{multirow}
\usepackage[table]{xcolor}
\usepackage{bm}
\usepackage{tablefootnote}
\usepackage{threeparttable}
\usepackage{color}
\usepackage{colortbl}
\usepackage{mdwmath,mdwtab}
\usepackage{autobreak}
\usepackage{amssymb}
\usepackage{multirow}
\usepackage{tabularx}
\usepackage{makecell}
\usepackage{threeparttable}
\usepackage{tablefootnote}
\usepackage[OT1]{fontenc}
\usepackage{float}
\newcolumntype{Y}{>{\centering\arraybackslash}X}
\usepackage{booktabs}
\usepackage{tabularx}
\usepackage[hang,flushmargin]{footmisc}
\usepackage{ragged2e}
\usepackage{fancyhdr}
\usepackage{indentfirst}
\usepackage{tablefootnote}
\usepackage{threeparttable}
\usepackage{color}
\usepackage{arydshln}
\usepackage{booktabs}
\usepackage{multirow}

\usepackage[numbers,sort&compress]{natbib}
\title{MPANet: Multi-Patch Attention For Infrared Small Target object Detection}
\name{Ao Wang, Wei Li~$^{*}$, Xin Wu, Zhanchao Huang, and Ran Tao
\thanks{This work was supported by the China Postdoctoral Science Foundation Funded Project No. 2021M690385, National Natural Science Foundation of China under Grant (62101045, 61571033, 91638201) and in part by the Aeronautical Science Foundation of China under Grant 20200051072001. (Corresponding Author: Wei Li; e-mail: liwei089@ieee.org).}}
\address{School of Information and Electronics, Beijing Institute of Technology, China}
\begin{document}
%
\maketitle
\begin{abstract}
	
Infrared small target detection (ISTD) has attracted widespread attention and been applied in various fields. Due to the small size of infrared targets and the noise interference from complex backgrounds, the performance of ISTD using convolutional neural networks (CNNs) is restricted. Moreover, the constriant that long-distance dependent features can not be encoded by the vanilla CNNs also impairs the robustness of capturing targets' shapes and locations in complex scenarios. To this end, a multi-patch attention network (MPANet) based on the axial-attention encoder and the multi-scale patch branch (MSPB) structure is proposed. Specially, an axial-attention-improved encoder architecture is designed to highlight the effective features of small targets and suppress background noises. Furthermore, the developed MSPB structure fuses the coarse-grained and fine-grained features from different semantic scales. Extensive experiments on the SIRST dataset show the superiority performance and effectiveness of the proposed MPANet compared to the state-of-the-art methods.
\end{abstract}
\section{Introduction}
In the past few years, dim infrared target detection has received more and more attention. It has been applied in various fields such as early warnings, missile tracking systems, and remote sensing. There are five difficulties in infrared small target detection, including small target size, complicated background , erratic target motion, unknown noise target motion, and a small quantity of single infrared targets.

For single-frame small target detection methods, they are divided into model-driven and data-driven methods. Several traditional methods are proposed to address the above challenges including filtering-based methods, local-contrast-based methods, and low-rank-based methods. 
However, the effectiveness of traditional approaches is very dependent on the prior knowledge of specific scenarios and manual algorithm hyperparameters. For this reason, some CNN-based methods have been applied in infrared small target detection in a data-driven manner.

\vspace{-0.2em}
Mostly, recent data-driven networks rely on a classification backbone to extract features. Dai et. al\cite{dai2021asymmetric} presented an asymmetric attention module to highlight small targets and public single frame infrared small target dataset. Dai et. al \cite{TGRS21ALCNet} developed an attention local contrast network. Dai et. al \cite{DBLP:journals/corr/abs-2106-00487} designed a dense nested interactive attention (DNIM) to achieve high level and low level feature fusion. Recently, Liu et. al \cite{DBLP:journals/corr/abs-2109-14379} is the first one to apply self- attention mechanism in infrared target detection. However, it is not sufficient to use vision transformer \cite{dosovitskiy2021image,9627165} architecture without considering the characteristics of the infrared small target itself. Hong et .al \cite{9170817} try to use the graph convolutional neural structure to overcome the Grid Sampling property of CNN. Wu et. al \cite{ 9598903} attempt to use multi-modal information fusion to get better detection results. Jeya Maria et. al \cite{valanarasu2021medical} adopted a local-global training strategy to improve the performance in medical segmentation. In addition, Lin et. al \cite{DBLP:journals/corr/abs-2106-06716} proposed a multi-scale transformer to capture long-range features of different semantic scales. In a multi-scale transformer, high performance is dependent on different scale patches, which is not sufficient for extracting global information.

To this end, a multi-patch attention network to resolve the problems above is designed. The contributions of this work are summarized as follows:

 1) A MPANet framework based on the axial-attention encoder and the multi-scale patch branch (MSPB) structure is proposed to highlight the small targets and suppress background without any classification backbones. In the designed MSPB, coarse-grained features extracted by the large-scale branch and fine-grained features extracted by the local branches are fused through the developed bottom-up structure.

  2) Extensive experiments on the public SIRST dataset demonstrate that the proposed MPANet achieves competitive results on the most of indicators compared to the existing model-driven methods and data-driven methods.

\section{PROPOSED MPANET FRAMEWORK}

\subsection{Axial-attention-based encoder}
\begin{figure}[!t]
    \centering
\epsfig{figure=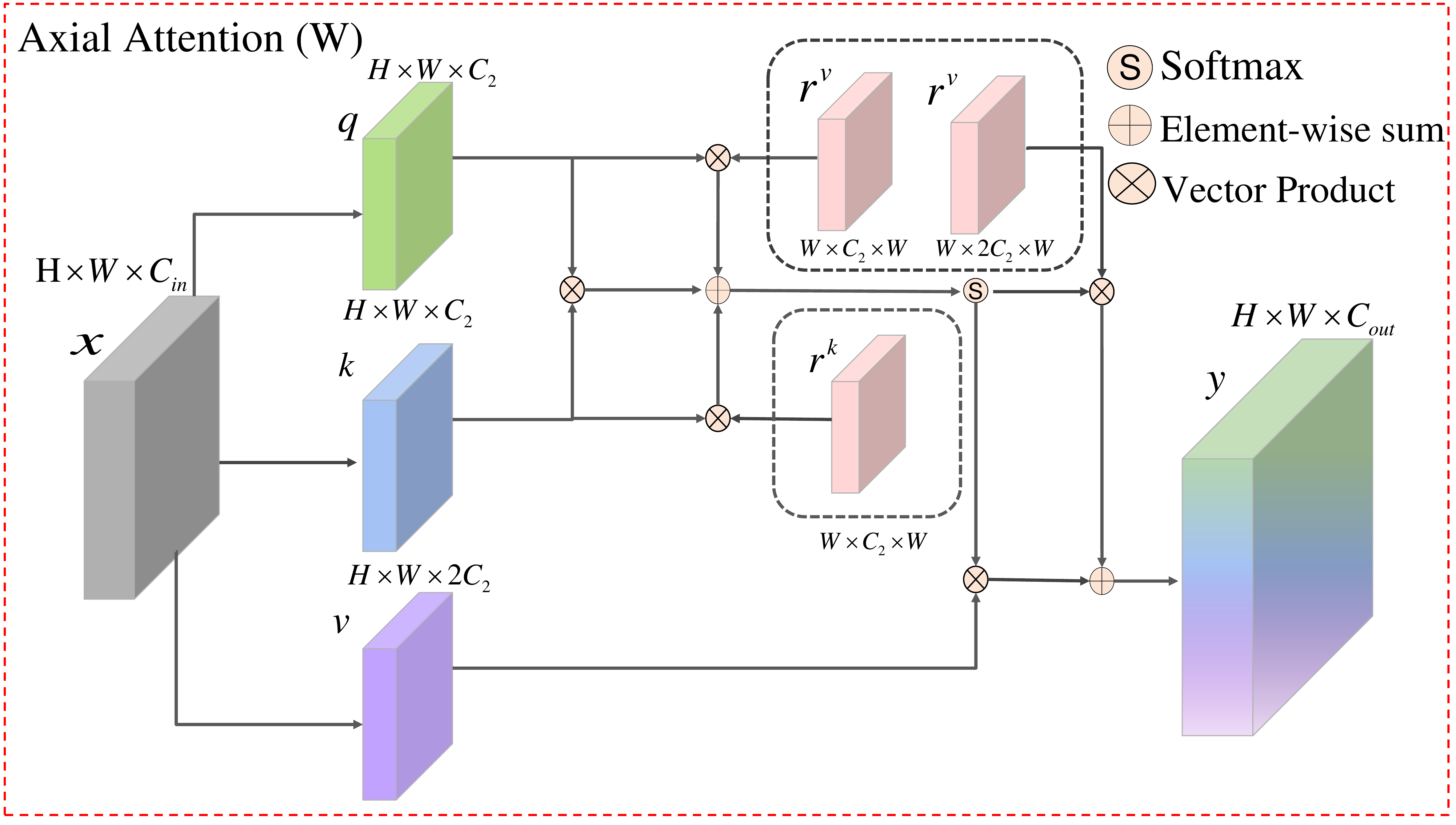,width=0.48\textwidth}
  \caption{The main structure of position-sensitivity axial attention, where $r^q,r^k,r^v \in \mathbb{R}^{\mathcal{W}\times \mathcal{W}} $ are the positional embedding for the width axis. $C_{in}$, $C_2$ and $C_{out}$ represent the number of channels of input, intermediate and output feature maps, respectively. When removing the positional embedding, position-sensitivity attention degenerates into non-local block. }
\end{figure}
To reduce original multi-head attention computation complexity and allow stacking self-attention layers within a larger or even global region, two-dimensional self-attention can be factorized into two one-dimensional self-attention.In addition, Wang et. al \cite{r1} combined the axial attention mechanism and use relative positional encodings for all queries, keys, and values. 
The projection matrices $\boldsymbol{W}_Q,\boldsymbol{W}_K,\boldsymbol{W}_V \in \mathbb{R}^{C_{in}\times C_{out}}$ are learnable during the training process.

An one-dimensional positive-sensitive self-attention on the special axis of an image is defined as the axial-attention layer. To improve the robustness of capturing targets’ shapes and
locations, axial-attention encoder is applied to enhance the ability of U-shaped structure to extract long-distance dependence,
\begin{small}
\begin{equation}
    y_{i,j}=\sum_{h=1}^{H}\sum_{w=1}^{\mathcal{W}}softmax(q_{ij}^T k_{hw})v_{hw}, 
    \label{eq:1}
\end{equation}
\begin{equation}
\begin{split}
    y_{i,j}=\sum_{w=1}^{\mathcal{W}}softmax_p(q^T_{ij}k_{iw}+q^T_{ij}r_{iw}^q\\ +k_{iw}^Tr_{iw}^k) 
    \times (v_{iw}+r_{iw}^v),
\end{split}
    \label{eq:2}
\end{equation}
\end{small}where $i\in\{1,\dots,H\}$ and $j\in\{1,\dots, \mathcal{W}\}$ represent any arbitrary location \cite{r1}. ${\boldsymbol q}=\boldsymbol{W}_Q{\boldsymbol x}$,  keys ${\boldsymbol k}=\boldsymbol{W}_k{\boldsymbol x}$ and values  ${\boldsymbol v}=\boldsymbol{W}_v{\boldsymbol x}$ are computed  from the input feature map $\boldsymbol{x}\in \mathbb{R}^{C_{in}\times H\times W}$.

The long-range affinities throughout the entire feature map can be captured by self-attention, where is advance to CNN-based methods.
However, non-local feature in  Eq.~\ref{eq:1} does not use any positional information with high computational complexity. For this, Eq.~\ref{eq:2} is calculated with relative position embeddings to capture global information. By means of employing two axial-attention layers consecutively
for the height-axis and width-axis, we can perform attention within different scale region and achieve an 
efficient self-attention module.

\subsection{The proposed MSPB structure}
 Due to low-resolution infrared images, the existing methods lose a lot of semantic information during multiple pooling processes. Therefore, it cannot be directly migrated to the small infrared target detection. In addition, the patch-wise training method makes the network unable to learn the mutual information or long-distance dependence between patches. To overcome the limitations mentioned above, an MSPB structure is adapted to perform corresponding aggregation operations from bottom to top.

\begin{figure*}[htb]
\begin{minipage}[b]{1.0\linewidth}
  \centering
\centerline{\epsfig{figure=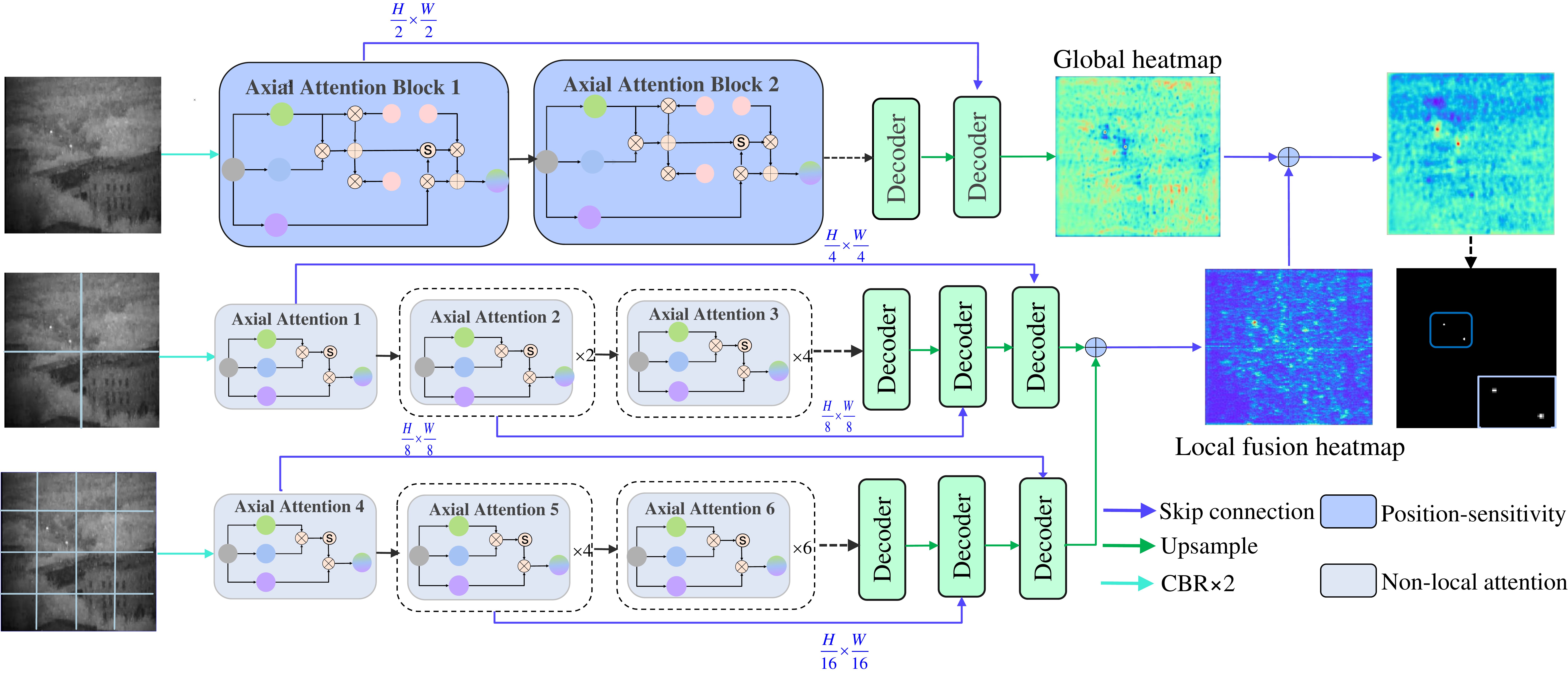,width=16cm}}
\vspace{-1em}
  \caption{The overview of the proposed MPANet. Given an input image, we split it into non-overlapping patches in three different scales.Among them, the branch of the original image resolution is regarded as global branch. In addition,we use position-sensitivity attention as the basic module of the encoder. The remaining branches are regarded as local branches. Correspondingly, non-local block become the basic module in local branches.The CBR in the legend stands for convolution, BatchNorm and Relu operation.}
  \label{fig:1}
\end{minipage}
\end{figure*}

 In the global branch, the position-sensitive attention block is used to capture the semantic and contour features of the foreground and background and the location information of the target. In the global heatmap in Fig.~\ref{fig:1}, it is observed that although the global branch highlight the object well, there are still a lot of noise and flashing circles in the background area.

  Conversely, the position information of patches in local branches is not less important than it in the global branch. Therefore, a non-local attention block is used in the two local branches to speed up the training process while reducing the number of parameters. Because the large-scale branch captures coarse-grained features better while the small-scale branch obtains the fine-grained feature better, a multi-branch architecture is designed to make  patches of different scales complement each other during feature extraction. Furthermore, local branches provide global branch more detailed target information. To a great extent, they simultaneously achieve strong suppression of the background. Similarly, in the local fusion heatmap in ~\ref{fig:1}, the background is effectively suppressed but the infrared small target is not very prominent. Finally, the structure improves the performance of detection through combining the high-level semantic features extracted by the global branch and the finer details acquired by the local branches. In the final heatmap in Fig.~\ref{fig:1}, the MSPB structure highlights the small
targets and suppresses background effectively.

\vspace{-1em}
\section{Experiments and Discussions}

\begin{figure*}[!tb]
	\vspace{-0.5em}
	\centering
	\epsfig{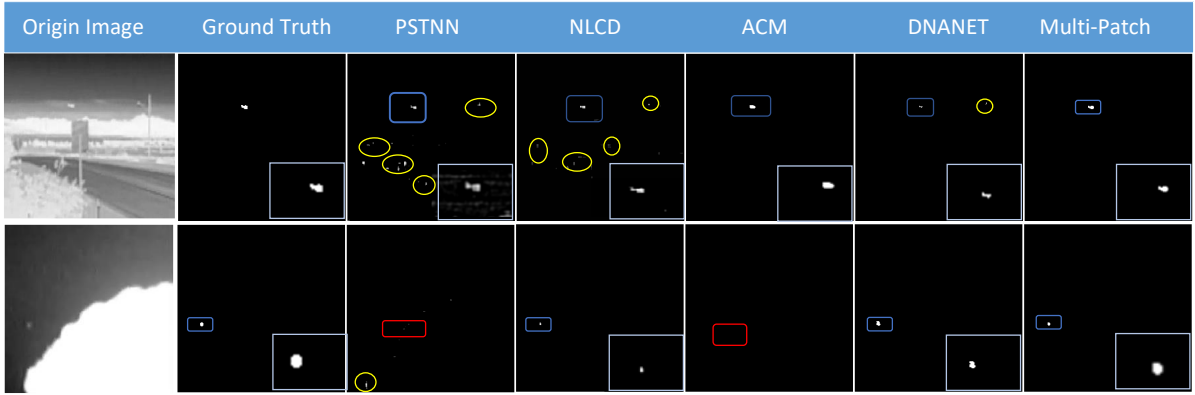}
	\vspace{-1em}
	\caption{Detection results predicted by some representative methods for SIRST dataset.The yellow ellipse represents the false alarm target,  the red square represents the missed target and blue bounding box means correct detection.}
	\label{fig:2}
	\vspace{-1em}
\end{figure*}

In this section, we adopt quantitative and quality evaluation metrics to evaluate different methods on the SIRST dataset \cite{dai2021asymmetric}, which is comprised of 427 images including 480 instances with high-quality annotations. SIRST includes short-wavelength, mid-wavelength, and 950 nm wavelength infrared images. In the experiments, the dataset is divided in a 3:1:1 ratio to form a train set, validation set and test set. For a fair comparison, all the input images and masks are cropped to fixed $256\times256$ pixels.
\vspace{-1em}
\subsection{Evaluation Metrics}
Considering that small target detection is regarded as a segmentation task, five quantitative indicators are used to evaluate various models. We use intersection over Union (IoU), normalized Intersection over Union (nIoU), and normalized $F_1$ score to evaluate shape description ability, 
\begin{small}
\begin{equation}
    nIoU = \frac{1}{N}\sum_i^N \frac{TP_i}{T_i+P_i-TP_i},
\end{equation}
\vspace{-0.5em}
\begin{equation}
     IoU  = \frac{TP}{T+P-FP},
\end{equation}
\end{small}where $T$, $P$, and $TP$ stand for label, predict result, and true
positive pixel numbers. In addition, $P_d$ and $F_a$ are used to evaluate localization ability, 
\begin{small}
\begin{equation}
    F_1 = \frac{1}{N}\sum_i^N\frac{{Precison_i}\times Recall_i}{Precision_i+Recall_i}
\end{equation}
\vspace{-0.5em}
\begin{equation}
    Pd = \frac{T_{correct}}{T_{All}}, 
    F_a = \frac{P_{false}}{P_{All}}
\end{equation}
\end{small}where $T_{correct}$ and $T_{All}$ represent the numbers of correctly
predicted targets and all targets. Correspondingly, $P_{false}$ and $P_{All}$ represent the numbers of falsely predicted pixels and all image pixels. If the centroid
derivation of the target is less than 3 pixels, the targets are considered as the correctly predicted. 
Otherwise, they are defined as false alarms. 
\subsection{Comparsion to the-state-of-art methods}
To demonstrate the superiority of the proposed methods, we compare several methods, including top-hat filters (Top-hat) \cite{tom1993morphology}, local contrast method (LCM) \cite{chen2013local}, infrared patch image (IPI)  \cite{gao2013infrared}, reweighted infrared patch-tensor (RIPT) \cite{dai2017reweighted}, multiscale grey difference weighted image entropy (MDGWIE) \cite{wang2019miss}, nonnegativity-constrained variational mode decomposition (NVMD) \cite{wang2017infrared}, partial sum of tensor nuclear norm (PSTNN) \cite{zhang2019infrared}, novel local contrast descriptor (NLCD) \cite{qin2019infrared}. Asymmetric contextual modulation (ACM) \cite{dai2021asymmetric}, miss detection vs false alarm generative adversarial network (MD vs FA) \cite{wang2019miss}, dense nested attention network (DNANet) \cite{DBLP:journals/corr/abs-2106-00487}. Especially, ACM is divided into U-Net and FPN architecture.

It is demonstrated that the proposed MPANet achieves the better performance on four indicators as listed in Table~\ref{tab:my-table}. The detection rate of DNANet is 1.5\% higher than that of the proposed MPANet, but the false alarm rate is increased by about 6 times. In the fine segmentation indicators, such as IoU and nIoU, the MPANet is 0.0268 and 0.0272 higher than those of the current CNN-based model. Furthermore, it also has a good performance on $F_1$ numerical indicators. In qualitative analysis, MPANet show excellent performance in Fig.~\ref{fig:2}. The network effectively reduces the probability of missed detection and false alarms. Although the mask labeling of the small infrared targets has marginal bias, multiple outputs indicate that the proposed MPANet effectively save the contour information and achieve the effect of smoothing edges.
\vspace{-1.5em}
\begin{table}[!htb]
\centering
\footnotesize
\caption{Quantitative comparsion of different Methods }
\label{tab:my-table}
\renewcommand\arraystretch{1.3}
\setlength{\tabcolsep}{1.2mm}{
\begin{tabular}{cccccc}
\toprule
Methods & IoU & nIoU & $ P_d {(10^{-2})} $  & $ F_a {(10^{-6})} $  & $ F1_{score} $   \\ \midrule 
LCM\cite{chen2013local} & 0.0027 & 0.1702 &75.229 &
13579.6 & 0.2517 \\ 
Tophat\cite{tom1993morphology}  & 0.2843   & 0.4073  & 89.908 &  412.51 &0.5394\\
IPI\cite{gao2013infrared}    & 0.4253    & 0.5014 & 96.3302  &  293.99 & 0.4719    \\ \
MGDWIE\cite{deng2016infrared}  & 0.1715     & 0.2093    & 89.2529           & 1024.1    & 0.3375 \\  
NLCD\cite{qin2019infrared}                     & 0.2333      & 0.2871   & 88.6238 & 625.96 & 0.4489  \\ 
RIPT\cite{dai2017reweighted}                          & 0.2348 & 0.3213 & 97.2477       & 293.99 & 0.4719 \\ 
NVMD \cite{wang2017infrared}                    & 0.2596 & 0.3903 & 88.9908  &869.39  & 0.5450 \\ 
PSTNN\cite{zhang2019infrared} & 0.4113   & 0.4917      & 91.0550 & 602.18 & 0.6412 \\ 
MDvsFA\cite{wang2019miss} & 0.5814  &0.6202 &74.0123
&439.15 & 0.6342\\
ACM FPN\cite{dai2021asymmetric} & 0.6359 & 0.6563 &96.3302 & 9.9359&0.7779 \\
ACM UNet\cite{dai2021asymmetric} & 0.6376 & 0.6549 & 97.2477 & 34.775 & 0.7825\\
DNANet\cite{DBLP:journals/corr/abs-2106-00487} & 0.7182 & 0.7237 &\textbf{99.5412} &  65.115 & 0.8302 \\ 
\hline
GloBal Branch & 0.7275 & 0.7194 & 95.2380 & 26.209 & \textbf{0.8387}\\
Local Global & 0.7367 &0.7397 & 98.0952 & 14.361 & 0.8269 \\

Multi-Patch & \textbf{0.7450} & \textbf{0.7509} & 97.1428 & \textbf{9.8733}&0.8282\\
\bottomrule
\end{tabular}}
\end{table}
\vspace{-2em}
\section{Conclusions}

In this paper, a MPANet based on self-attention mechanism has been proposed for infrared small target detection. We use lightweight U-shape-like architecture and replace the convolution in the encoder with axial attention block. Using MSPB structure, the network realizes the bottom-up fusion from low-scale patch
semantics to global image semantic information, which can effectively
extract the coarse and fine-grained feature representations
of different semantic scales. The experiment results on the public SIRST dataset suggest the excellent performance on fine segmentation and target localization ability.

\section{REFERENCES}
\vspace{-3em}
\scriptsize%
\bibliographystyle{IEEE.bst}
\bibliography{DKED.bib}

\end{document}